\documentclass[11pt]{article}
\usepackage{acl2014}
\usepackage{times}
\usepackage{url}
\usepackage{latexsym}
\usepackage{paralist}
\usepackage[utf8]{inputenc}
\usepackage{amsmath}

\title{Chinese Restaurant Process for cognate clustering: A threshold free approach}

\author{Taraka Rama \\
  Institute of Linguistics \\
  University of Tübingen \\
  Germany \\
  {\tt taraka-rama.kasicheyanula@uni-tuebingen.de}}

\date{}

\begin{document}
\maketitle
\begin{abstract}
In this paper, we introduce a threshold free approach, motivated from Chinese Restaurant Process, 
for the purpose of cognate clustering. We show that our approach yields similar results to 
a linguistically motivated cognate clustering system known as LexStat. Our Chinese Restaurant 
Process system is fast and does not require any threshold and can be applied to any language family 
of the world.
\end{abstract}

\section{Introduction}
Identification of cognates is an important task while establishing genetic relations between 
languages that 
are hypothesized to have descended from a single language in the past. For instance, English 
\emph{hound} and 
German \emph{Hund} are cognates which can be traced back to Proto-Germanic stage.
Highly accurate automatic identification of cognates is desired for reducing the effort 
required in analyzing large language families such as Indo-European \cite{bouckaert2012mapping} and 
Austronesian \cite{greenhill2009austronesian} can take up decades of effort, when performed by 
hand. A automatic cognate identification system can be helpful for historical linguists to analyze 
supposedly related language families and also fasten up the making of cognate databases which can 
be then be analyzed using Bayesian phylogenetic methods \cite{Atkinson:06}.

In this paper, we work with Swadesh word lists of multiple language groups and attempt to cluster 
related words together using a non-parametric process known as Chinese Restaurant Process 
\cite{gershman2012tutorial}. We use the sound similarity matrix trained in an unsupervised fashion 
\cite{jager2013phylogenetic} for the purpose of computing similarity between two words. The CRP 
based algorithm is similar to the CRP variant of the K-means algorithm introduced by 
\newcite{kulis2011revisiting}. Our CRP algorithm does not require any threshold and only has a 
single hyperparameter known as $\alpha$ which allows new clusters to be formed without the 
requirement of threshold or the number of clusters to be known beforehand.

Previous work by \newcite{list-lopez-bapteste:2016:P16-2} and 
\newcite{hauer-kondrak:2011:IJCNLP-2011} employ a hand crafted or a machine learned word similarity 
measure to compute pair-wise distances between words. The pair-wise distance matrix is then 
supplied to a clustering algorithm such as average linkage clustering \cite{Manning:99} for 
inferring a tree structure of the words. The average linkage clustering algorithm is an 
agglomerative algorithm that merges individual clusters until a single cluster is left. The 
clustering process can be interrupted if the average similarity between two clusters falls below a 
predetermined threshold.

The agglomerative algorithm is simple and usually yields reasonable results across various language 
families \cite{list:2012:LINGVIS2012}. However, the method suffers from a major drawback that the 
threshold needs to be known beforehand for achieving high accuracy. In a recent paper, 
\newcite{list-lopez-bapteste:2016:P16-2} use a clustering algorithm known as InfoMap for the 
purpose of clustering cognates in Sino-Tibetan language groups. The InfoMap algorithm also requires 
a threshold for finding cognates. The authors find that the algorithm works well if the threshold 
is adjusted across language groups. In this paper, we compare our system against the LexStat system 
and show that our system yields comparable results.

The structure of the paper is as followed. We define the cognate clustering problem in section 
\ref{sec:cogclust}. In section \ref{sec:seqalign}, we describe the 
string alignment algorithm for the purpose of computing similarity between two strings. We describe 
our CRP algorithm in section \ref{sec:crp}. We describe the evaluation of our experiments 
and datasets in section \ref{sec:exps}. We present the results of our experiments and discuss them 
in section \ref{sec:results}. We conclude the paper in section \ref{sec:concl}.



\section{Cognate clustering}\label{sec:cogclust}

The phylogenetic inference methods require cognate judgments which are only available for a small 
number of well-studied language families such as Indo-European and Austronesian. For instance, the 
ASJP database \cite{brown2013sound}\footnote{\url{asjp.clld.org}}
provides Swadesh word lists (of $40$ length that are supposedly important for identifying genetic 
relationships between languages) 
transcribed in a uniform format for more than $60\%$ of the world's languages.\footnote{However, 
the cognacy 
judgments are only available for a subset of language families.} An example of such a word list is 
given below:
\begin{table}[!ht]
  \centering
\small
    \begin{tabular}{|l|cccc|}
   \hline
       & ALL & AND & ANIMAL & $\ldots$\\
          \hline
    English & ol & End & Enim3l & $\ldots$\\
    German & al3 & unt & tia & $\ldots$\\
    French & tu & e  & animal & $\ldots$\\
    Spanish & to8o & i  & animal & $\ldots$\\
    Swedish & ala & ok & y3r & $\ldots$\\
       \hline
    \end{tabular}%
    \caption{Example of a word list, in ASJP transcription for five languages belonging to Germanic 
(English, German, and 
Swedish) and Romance (Spanish and French) subfamilies.}
\label{tab:Datasample}
\end{table}

The task at hand is to automatically cluster words according to genealogical relationship. This is 
achieved by computing similarities between all the word pairs belonging to a meaning and then 
supplying the resulting distance matrix as an input to a clustering algorithm. The clustering 
algorithm groups the words into clusters by optimizing a similarity criterion. The similarity 
between a word pair 
can be computed using supervised approaches \cite{hauer-kondrak:2011:IJCNLP-2011} or by using 
sequence alignment 
algorithms such as Needleman-Wunsch \cite{needleman1970general} or Levenshtein distance 
\cite{levenshtein1966binary}. An example of a pairwise distance matrix for meaning ``all'' is shown 
in table \ref{tab:sampledist}.

\begin{table}[!ht]
\small
  \centering
    \begin{tabular}{|l|c|c|c|c|c|}
\hline
       & ol & al3 & tu & to8o & ala \\
\hline
    ol &    & 0.28 & 0.99 & 0.99 & 0.4 \\
\hline
    al3 & 0.28 &    & 0.94 & 0.99 & 0.01 \\
\hline
    tu & 0.99 & 0.94 &    & 0.55 & 0.99 \\
\hline
    to8o & 0.99 & 0.99 & 0.55 &    & 0.99 \\
\hline
    ala & 0.4 & 0.01 & 0.99 & 0.99 &  \\
\hline
    \end{tabular}%
    \caption{An example of a pairwise distance matrix between all the words 
for meaning ``all''.}
\label{tab:sampledist}
\end{table}%

\section{Sequence alignment}\label{sec:seqalign}

The Needleman-Wunsch algorithm is the similarity counterpart of the Levenshtein distance. The 
Needleman-Wunsch algorithm maximizes similarity whereas Levenshtein distance minimizes the 
distance. 
In the Needleman-Wunsch algorithm, a character or sound segment match increases the similarity by 
$1$ 
and a character mismatch has a 
weight of $-1$. In contrast to Levenshtein distance which treats insertion, deletion, and 
substitution equally, the Needleman-Wunsch algorithm introduces a gap opening (deletion 
operation) penalty parameter that has to be learned separately. A second parameter known as gap 
extension penalty has lesser penalty than the gap opening parameter and models the fact that 
deletions occur in chunks \cite{jager2013phylogenetic}. 

The (vanilla) Needleman-Wunsch algorithm is 
not sensitive to segment pairs and a realistic algorithm should assign high similarity between 
sound correspondences such as /s/ $\sim$ /h/ than the sound pair /p/ $\sim$ /r/.

In dialectology \cite{wieling2015advances}, similarity between two segments is estimated using PMI. 
The PMI score of two 
sounds 
$i$ and $j$ is defined as followed:
\begin{equation}\label{eq:PMI} 
\textnormal{PMI}(i,j) = \log (\frac{p(i,j)}{q(i)\cdot q(j)})
\end{equation}
where, $p(i,j)$ is the relative frequency of $i, j$ occurring at the same position in th aligned 
word pairs whereas, $q(.)$ is the relative frequency of a sound in the whole word list. A positive 
PMI value indicates that a segment pair cooccurs together whereas, a negative PMI value indicates 
lack of cooccurrence. This can be interpreted as a strength of relatedness between two segments.

In this paper, we use the PMI matrix (of ASJP sound segments) inferred by 
\newcite{jager2013phylogenetic} for computing the similarity between a word pair. 
\newcite{jager2013phylogenetic} shows that the PMI matrix shows positive weights for sound pairs 
such as /p/ $\sim$ /b/, /t/ $\sim$ /d/, and /s/ $\sim$ /h/. 

\section{CRP}\label{sec:crp}
In this section, we describe the CRP algorithm and motivate its suitability for cognate 
clustering. Given a meaning $M$ and the word similarity matrix $S$ of 
dimensions $N \times N$, the CRP algorithm works as follows. The CRP outputs 
$K$ clusters and the clustering $l_1,\dots l_K$.
\begin{enumerate}
 \item Initially, assign a word $w_n$ to $l_n$ where $K=N$.
 \item Repeat until convergence:
 \begin{itemize}
  \item For each word $w_n$:
    \begin{itemize}
      \item Remove $w_n$ from its cluster.
     \item Compute $s_{nk}$ the average similarity of $w_n$ to all words in 
cluster $k$.
     \item If $\underset{k}{\arg\max}$ $s_{nk} < \alpha$ assign $w_n$ to a new 
cluster.
     \item Else, assign $w_n$ to the cluster $k$ where $k 
=\underset{k}{\arg\max}$ $s_{nk}$.
    \end{itemize}

 \end{itemize}

\end{enumerate}

The current algorithm uses the criterion of average similarity to assign a word to a cluster. A 
word is assigned to the cluster with which it exhibits the highest average similarity. The 
intuition behind this decision is that the word should, on an average, be similar to the rest of 
the words in a cluster. \footnote{The average similarity criterion can be modified to the maximum 
similarity criterion. This is commonly known as single linkage clustering \cite{Manning:99}.}

The magnitude of the $\alpha$ parameter determines the number of new clusters. 
A value of $0.01$ is sufficient for the purpose of forming new 
clusters. The word similarity is always non-negative and we use a ReLU 
transformation ($max(0,x)$) that transforms negative similarity scores to $0$. The CRP algorithm 
identifies cognate clusters of uneven sizes and can also form singleton clusters due to the simple 
initialization. In our experiments, we find that three full scans of the data are sufficient for the 
algorithm to reach a local maximum.
















\section{Experiments}\label{sec:exps}
\textbf{Baseline} We use a vanilla Needleman-Wunsch with a gap opening penalty of $-1$ and a gap 
extension penalty of $-0.5$ as the baseline in our experiments. 

\textbf{LexStat} LexStat \cite{list:2012:LINGVIS2012} is a system offering state-of-the-art 
alignment 
algorithms for aligning word pairs and clustering them into cognate sets. The LexStat system weighs 
matches between sounds using a handcrafted segment similarity matrix that is informed by historical 
linguistic literature.

\subsection{Evaluation}
We evaluate the results of clustering analysis using B-cubed F-score \cite{amigo2009comparison}. 
The B-cubed scores are defined for each individual item as followed. The precision 
for an item is defined as the ratio between the number of cognates in its cluster to the total 
number of items in its cluster. The recall for an item is defined as the ratio between the number 
of 
cognates in its cluster to the total number of expert labeled cognates. The B-cubed precision and 
recall are defined as the average of the items' precision and recall across all the clusters. 
Finally, the B-cubed F-score for a meaning, is computed as the harmonic mean of the average items' 
precision and recall. The B-cubed F-score for the whole dataset is given as the average of the 
B-cubed F-scores across all the meanings.

Both \newcite{hauer-kondrak:2011:IJCNLP-2011} and 
\newcite{list-lopez-bapteste:2016:P16-2} use B-cubed F-scores to test their cognate clustering 
systems.

\subsection{Datasets}
\textbf{IELex database} The Indo-European Lexical database was created by
\newcite{dyen1992indoeuropean} and curated by Michael Dunn. The IELex database is not 
transcribed in uniform IPA and retains many forms transcribed in 
the Romanized IPA format of \newcite{dyen1992indoeuropean}. We cleaned the IELex database of any 
non-IPA-like transcriptions and converted the cleaned subset of the database into ASJP format. The 
cleaned subset has $52$ languages and $210$ meanings.

\textbf{Austronesian vocabulary database} The Austronesian Vocabulary 
Database (ABVD) \cite{greenhill2009austronesian} has word lists for $210$ Swadesh concepts and 
$378$ 
languages.\footnote{\url{http://language.psy.auckland.ac.nz/austronesian/}} The database 
does not have transcriptions in a uniform IPA format. We 
removed all symbols that do not appear in the standard IPA and converted the lexical items to ASJP 
format. For comparison purpose, we use randomly selected $100$ languages' dataset in this 
paper.\footnote{LexStat takes many hours to run on a dataset of $100$ languages.}

\textbf{Short word lists with cognacy judgments} \newcite{wichmann2013languages} and 
\newcite{List2014d} compiled cognacy wordlists for subsets of families from 
various scholarly sources such as comparative handbooks and historical linguistics' articles. 
The details of this compilation is given below. For each dataset, we give the number of 
languages/the number of meanings in parantheses. 

\begin{compactitem}
 \item \newcite{wichmann2013languages}: Afrasian (21/40), Mayan (30/100), Mixe-Zoque (10/100), 
Mon-Khmer (16/100).
\item \newcite{List2014d}: ObUgrian (21/110; Hungarian excluded from Ugric sub-family).
\end{compactitem}


\section{Results}\label{sec:results}

The B-cubed F-scores of different systems are shown in table \ref{tab:results}. The CRP based PMI 
system performs better than LexStat on four datasets. The CRP algorithm performs slightly worse 
than the LexStat system on Austronesian and Indo-European language families by two points. The 
LexStat performs better than the PMI-CRP system only on the Ugric languages dataset. The LexStat 
system's clustering threshold has been tuned on many smaller datasets whereas, the PMI-CRP does not 
require any tuning of the threshold and comes closer or performs better than the LexStat system. We 
also provide the average of B-cubed F-scores across different datasets. The results show that the 
PMI-CRP system is close to the performance of the LexStat system.

\begin{table}[!ht]
\centering
\small
 \begin{tabular}{l|ccc}
\hline
  Dataset&LexStat&NW-CRP&PMI-CRP\\\hline
Afrasian&76.54&78.2&\textbf{81.22}\\
Austronesian&\textbf{74.9}&71.89&72.39\\
Mayan&78.6&80.38&\textbf{80.75}\\
Mixe-Zoque&91.45&88.61&\textbf{92.35}\\
Indo-European&\textbf{77.56}&67.2&74.89\\
Mon-Khmer&80.49&78.11&\textbf{81.69}\\
ObUgrian&\textbf{92.19}&73.4&86.78\\\hline
Average& 81.68&76.83 & 81.44\\\hline
 \end{tabular}
\caption{Average B-cubed F-scores of different systems. The suffix ``-CRP'' stands for the CRP 
algorithm applied to Needleman-Wunsch (NW) and PMI word similarity methods.}
\label{tab:results}


\end{table}

\subsection{Match between predicted and obtained clusters}
We examine the match between the number of predicted clusters and the number of true 
clusters for the PMI-CRP system across meanings. We report the correlations in table 
\ref{tab:correl}. The correlations suggest that the number of predicted clusters correlate highly 
with the true number of clusters across datasets.

\begin{table}[!ht]
\centering
\small
 \begin{tabular}{l|c}
\hline
Dataset & PMI-CRP\\\hline
Afrasian & 72.69\\
Austronesian & 70.66\\
Mayan & 81.35\\
Mixe-Zoque & 88.94\\
Indo-European & 76.41\\
Mon-Khmer & 82.11\\
Ugric & 73.82\\\hline

 \end{tabular}
\caption{Pearson's correlation between the true number of clusters and the number of predicted 
clusters across language families.}
\label{tab:correl}

\end{table}
\subsection{Error analysis}
In the case of Indo-European, the PMI-CRP system fails to group all the reflexes for the meaning 
``five'', ``fingernail'', ``three'', ``two'', and ``name'' into a single cognate cluster. The 
reason for this behaviour is the extensive phonological change that affected cognates across the 
daughter subgroups. The LexStat system also shows similar behaviour when the true number of cognate 
clusters is $1$.

\section{Conclusion}\label{sec:concl}
In this paper, we introduced a CRP based clustering algorithm that is threshold free. The 
program takes less than two minutes for clustering a large dataset of 100 languages such as 
Austronesian. We tested the algorithm on a wide range of language families and showed the algorithm 
yields close or better results than LexStat. Based on the results, we claim that the algorithm can 
be useful for the comparative linguists to analyze putative language relations at a quick pace.

The main limitation of the algorithm is that it fails to retrieve clusters for meanings such as 
``what'', ``who'', and ``we'' (in Indo-European) which show high phonological divergence. In 
comparison, even LexStat makes mistakes when clustering these meanings. Whenever the reflexes show 
similar word forms, in the case of Mayan (meanings: ``water'' and ``die''), the algorithm groups 
all the reflexes into a single cluster without any error.

As part of future work, we plan to use the CRP algorithm for clustering meanings across different 
language families available in the ASJP database and then supply the cognate clusters to a Bayesian 
phylogenetic inference software such as MrBayes \cite{ronquist2003mrbayes} for inferring Bayesian 
trees for the languages of the world.


\bibliography{eacl2017}
\bibliographystyle{acl}

\end{document}